\documentclass[]{article}

\usepackage{authblk}
\usepackage{booktabs}
\usepackage{listings}
\usepackage{graphicx}
\usepackage{xurl}
\usepackage[numbers]{natbib}
\AtBeginDocument{\bibsep=0pt}
\title{Setting the AI Agenda -- Evidence from Sweden in the ChatGPT Era}

\author[1]{Bastiaan Bruinsma\footnote{Corresponding author. All authors contributed equally to the paper}}
\author[2]{Annika Fredén}
\author[2]{Kajsa Hansson}
\author[1]{Moa Johansson}
\author[3]{Pasko Kisić-Merino}
\author[1]{Denitsa Saynova}

\affil[1]{Chalmers University of Technology}
\affil[2]{Lund University}
\affil[3]{Karlstad University}

\begin{document}

\maketitle
 
\begin{abstract}
This paper examines the development of the Artificial Intelligence (AI) meta-debate in Sweden before and after the release of ChatGPT. From the perspective of agenda-setting theory, we propose that it is an elite outside of party politics that is leading the debate -- i.e. that the politicians are relatively silent when it comes to this rapid development. We also suggest that the debate has become more substantive and risk-oriented in recent years. To investigate this claim, we draw on an original dataset of elite-level documents from the early 2010s to the present, using op-eds published in a number of leading Swedish newspapers. By conducting a qualitative content analysis of these materials, our preliminary findings lend support to the expectation that an academic, rather than a political elite is steering the debate. 
\end{abstract}
\vspace{3mm}

\section{Introduction}

Since the release of ChatGPT in November 2022, concerns about the potential promises and pitfalls of artificial intelligence (AI) have raised considerable attention. On the one hand, AI is seen as a potential solution to several emerging challenges, such as health crises, food security, and climate change. On the other hand, the debate surrounding the development of AI seems to revolve mainly around the risks associated with it. In March 2023, the Future of Life Institute published an open letter, signed by both AI researchers and industry CEOs, calling for a six-month ``pause'' on the development of large language models\footnote{\url{https://futureoflife.org/open-letter/pause-giant-ai-experiments/}}. Since then, much media attention has focused on \emph{long-term} risks and the fears of AI leading to human extinction, with journalists and authors such as Ezra Klein\footnote{The New York Times, 12-03-2023} and Yuval Noah Harari\footnote{The New York Times, 24-03-2023} claiming that a critical mass of AI experts believe that AI may pose existential risks to the future. One of the most influential of these  is Eliezer Yudkowsky, who has argued for shutting down AI development\footnote{The Times, 29-03-2023}.

However, this focus on long-term risk has been strongly criticised by other AI researchers, such as Timnit Gebru \cite{Bender2021}. As one of the authors of a statement published by the DAIR Institute\footnote{\url{https://www.dair-institute.org/blog/letter-statement-March2023/}} in response to the ``pause letter'', Gerbu and her co-authors called instead for a focus on concrete \emph{short-term} risks. Instead of the existential focus of the long-term risks, these risks concern the immediate consequences of AI, such as that on job displacement and the spread of political misinformation and disinformation \citep{AcemogluJohnson2023, WorldEconomicForum2023a}, as well as concerns about many aspects of the (un)fairness and bias inherent to AI systems \citep{mehrabi2022, luccioni23,bolukbasi2016,pmlr-v81-buolamwini18a} (see also \citet{taxonomyLLMRisk} for a discussion of the risks of large language models specifically). This perspective is also promoted by influential journals such as Nature \citep{Nature2023}. In response to these discussions and debates on AI risk, the European Commission has moved swiftly to introduce legislative measures aimed at regulating the use of AI and assessing the associated risks \citep{EU2024a}\footnote{See also: \url{https://artificialintelligenceact.eu/}}.

Despite being high on the agenda, there are relatively few articles that consider AI from the perspective of the meta-discussion (for an exception see \citet{Nguyen2023a}). Rather, research on AI tends to take a normative stance, focusing on ``responsible'' AI (see, for example \citet{Dignum2019a} and \citet{Hedlund2022a}), issues of transparency \citep{LarssonHeintz2020} or the more technical aspects of AI performance \citep{Heseltine2024, Zhang2023, Gilardi2023}. Instead of this, we want to take a bird's eye view of the AI debate by examining developments in a highly digitised member of the European Union: Sweden. 

Previous research suggests that national media are often the most important source of public agenda setting, at least in countries such as the UK and Canada \citep{Langer2021a, Dandurandetal2023}. \citet{Langer2021a} suggests that there is a reciprocal relationship between elite politics and the agenda-setting media, so that politics and editorial gate-keeping are intertwined.  In this particular case, a tentative hypothesis is that the academic elite is more active in constituting the debate because because it takes a certain level of confidence and knowledge of an issue to pass through the eye of the editorial needle. If an academic elite dominates the AI issue, this would represent a shift in a party-oriented system such as Sweden, where politicians have traditionally played a major role in shaping people's opinions \citep{Holmberg1997a}.

The material in our study comes from argumentative texts (opinion pieces) by elite representatives published in leading Swedish newspapers. As the texts consist mainly of arguments with important information between the lines, we conduct a qualitative content analysis. This is carried out by three of the authors, each of whom is a native speaker with a degree in either social sciences or computer science. During the coding process, we focus on categorising the core content of the texts and classifying whether an article emphasises short-term or long-term risk, or does not mention risk at all. For our purposes here, we define short-term as authors writing about concrete risks that could occur in the near future related to the implementation of AI, whereas long-term risks are those that refer to hypothetical or existential risks instead.

With our approach, we contribute to the meta-debate on the basis of empirical evidence, a combination that is largely lacking in current social science studies (where \citet{AcemogluJohnson2023, Dandurandetal2023} are exceptions). Our analysis supports the claim that the debate is being led by an academic rather than a political elite. We also see a trend towards a greater emphasis on short-term risks from 2022 onwards. We discuss the implications of this and relate the debate in Sweden to the international debate on agenda-setting in the ChatGPT era.

\section{Agenda-Setting through Institutions}

A starting point for our study is that what is said in the leading newspapers remains an important source of opinion formation for citizens. A seminal paper by \citet{McCombs1972a} found that the news media played an important role in telling citizens what to think about, in effect setting the agenda. On the other hand, the media's influence, from their perspective, was more limited in terms of influencing what line of argument or thought citizens should have. Rather, it acts as a headlamp, telling people what to think about. More recent work has even questioned the agenda-setting role of traditional media in this way. One argument is that today's plethora of media channels and sources makes it difficult to develop common messages or a ``national agenda'', as people tend to self-select into channels that support their previous interests or views. In the case of the United States, Bennet and Iyengar concluded back in \citeyear{Bennett2008a} that these behaviours can make traditional media agenda-setting minimal. 

More recent evidence from Sweden, the case for our study, shows that the media continues to play an important role. \citet{Shehata2013a} examine the causal effect of coverage in the traditional news media on perceived importance by the public. They find that issues that received more coverage in the traditional news media tended to be perceived as more important by the public over time. In contrast, no such effects were documented for issues that were low on the media agenda. \citet{DjerfPierre2024a} address how media salience from different sources (news media and alternative media) uniquely affects the relative strength or weakness of ``sociotropic'' beliefs. They argue that while alternative and social media have become important elements in agenda setting and issue salience in Sweden their findings point to a continued important role for traditional news media. Given that coverage in the traditional news media has an impact on the perceived salience to the general public in Sweden \citep{Shehata2013a}, the number of articles on AI as such is important. This supports the idea of \citet{McCombs1972a} that the more coverage there is, the more people will pay attention to it.

The extent to which the national agenda is top-down or bottom-up driven also depend on the party system. Some research from the US suggests that US state legislators are very responsive to public opinion from below \citep{Barbera2019a}. In the US case, the perspective of the people or party supporters is thus important relative to the party elite. \citet{Holmberg1997a} contrasts the United States with the European multi-party tradition with strong party cohesion and concludes that a top-down perspective on opinion formation is much more relevant and prevalent in the case of European countries than in the United States. Looking at the case of Sweden, Holmberg finds that, for example, on the issue of computers and robotics (which developed in the 1980s and 1990s), the Swedish electorate followed the opinions of party representatives rather than vice versa. This lends some support to the idea that the political elite influenced the public on highly technical issues during this period, rather than vice versa. In addition, recent research from the United Kingdom context suggests that the influence of the media should not be underestimated. \citet{Langer2021a} find that in the UK context, the national media  tend to lead a national conversation on an issue. These authors discuss a reciprocal relationship between the political elite and the media. 

The study of agenda-setting dynamics in the public sphere has mostly been advocated for the cases of the United States and the United Kingdom, but recent scholarship has turned its attention to other contexts. Examining the South Korean context, \citet{Zhang2024a} investigates how social media have affected the prevailing ``rules'' of agenda setting and argues for the consideration of a ``multi-participant agenda setting'' dynamics. The authors consider agenda setting in relation to the interplay between bots, news media and the public in the context of presidential elections and the salience of key issues in the public sphere. In this context agenda setting appears to partially shift away from the news media over time, following both the ``bot agenda'' and the ``public agenda''. 

\section{The AI Issue}

Agenda setting scholarship has also begun to explore the dimension of technology and AI in the context of techno-political ``change'' in modern societies, although research to date is limited. For example, \citet{Nguyen2023a} addresses the context of public debates about AI and its more sensitive dimensions, such as privacy, and how these technologically complex issues are communicated to the general public by Anglophone-Western media. Using a mixed-methods approach to these debates, the author finds that the extent and focus of media attention is directly related to the actual spread of AI across different ``societal domains'' - e.g. politics, finance and healthcare.

Using topic modelling, Nguyen finds that over time, these associations have become more focused on the techno-social implications of AI in areas such as international politics, security and finance over time. Indeed, with the caveat that the sample is overwhelmingly US-based, Nguyen argues that this signals an increasing politicisation of AI. This link between AI and politics in the context of $2010--2022$ is further explored through the identification of four ``risk'' categories related to cybercrime/cyberwar, information disorder (mis- and disinformation), surveillance, and data bias. Specifically, Nguyen looks at the increase in ``explicit'' data risk references in the media (news articles) over time and finds that a total of 47 percent of AI-centric articles explicitly address risks, implying a shift towards an overall more attentive media coverage. However, by the end of the data collection period ($2020--2021$), mentions of risk decreased in Nguyen's sample. 

From a more critical perspective, \citet{Dandurandetal2023} show that the context can matter for the type of debate and information that news media disseminate. They examine the news framing of AI in the Canadian context, where they argue that mainstream media are reluctant to cover controversies in AI development. The authors claim that at the heart of this multi-stakeholder process of ``freezing out'' controversies is the goal of promoting Canada’s artificial intelligence ecosystem as a national imperative, where Canada has a reputation for leading the AI development. They also point to the weakness of journalists vis-à-vis tech experts on AI, which signals a weakening of democratic checks and balances. 

In a very recent piece, \citet{xian2024landscape} look specifically at the coverage and characteristics of English-language (mostly US-based) news articles on generative AI such as ChatGPT. The authors address the ability of these media to shape perceptions of generative AI technologies over time. To address these issues, the authors use a mixed-methods approach that combines topic modelling, qualitative coding, and sentiment analysis. Similar to \citet{Nguyen2023a}, the topic modelling shows that news coverage depends on ``hype'' or ``spikes'' corresponding to major AI breakthroughs and policy debates around generative AI. The authors find that key topics such as ``regulation and security'' correspond precisely to these temporal spikes. 

However, this topical distribution differs between national contexts. For example, US-based coverage gives more attention to the topic of ``business'' and ``technological development'', while Indian coverage is significantly more focused on ``corporate technological development'', and UK and Australian coverage focuses on ``regulation and security'' and ``technology development''. In addition, business-oriented articles tend to frame generative AI in a more positive or neutral light, while security and regulation articles (which increasingly focus on AI governance) tend to be more neutral or negative in their framing. Furthermore, the authors find that US-based articles tend to be more positive overall in their framing of generative AI technologies than the non-US Anglophone sample, but mostly in the case of specialised and local news outlets.

Our work is related to the perspective of \citet{Nguyen2023a} and \citet{xian2024landscape}, with the difference that we focus explicitly on argumentative debate articles in mainstream newspapers. In contrast to Nguyen, we cover a period both before and after the release of ChatGPT and focus on a case that has not been the focus of similar studies so far: Sweden.

\section{The Swedish Case}

Sweden is a parliamentary system where political parties have traditionally played an important role in organising people and people's opinions (see \citet{Holmberg1997a}). Since 2022, the government has consisted of a centre-right minority, although the Social Democrats are still by far the largest party. At the national government level, Sweden's current AI focus is on education, democracy and the media. The newly established Mediemyndigheten, initiated by the government, aims to increase citizens' knowledge about AI and disinformation\footnote{\url{https://www.regeringen.se/pressmeddelanden/2024/03/mediemyndigheten-ges-i-uppdrag-att-genomfora-nationell-satsning-for-starkt-medie--och-informationskunnighet-inom-ai-driven-desinformation/}}. Another stakeholder group in AI is business, which has received less attention in the public debate.

As a member of the European Union since 1995, Sweden is one of the countries that will adopt a new AI law initiated by the EU Commission and passed by the European Parliament in March 2024. This law focuses on restricting how AI tools can be used in practice, depending on whether the risk to humans is considered minimal, limited, high or unacceptable. While the Swedish parliament has been relatively silent on AI issues, the supranational level of the EU is thus at the forefront of regulation, as is the case in some other transnational areas such as climate change policy. Influences from the US and the continent have also spread to academics and elite debaters. The US debate \citep{AcemogluJohnson2023}, where AI is being developed as a potential threat to workers and/or as a real-time replacement for humans in the public sector, has begun to spread to Sweden. Sweden also has a vocal academic in the AI debate, physics professor Max Tegmark, who is sometimes recruited as an expert in the national media, and who took the initiative to the aforementioned ``pause letter''. Another leading scholar in the national and international debate is professor of mathematical statistics Olle Häggström, based at Chalmers University of Technology, who has recently become more sceptical to the rapid development of AI \citep{Haggstrom2023}.

With this in mind, our overarching research question is:

\begin{itemize}
    \item[Q: ] How has the Swedish debate on AI evolved? 
\end{itemize}

From our preliminary understanding of the context, we have the following expectations: 

\begin{itemize}
    \item[H1: ] The debate is growing.
    \item[H2: ] Arguments about risks are increasing
    \item[H3: ] The risk debate is being led by an academic and industrial elite rather than a political one.
\end{itemize}

\section{Method and Data}

Given the nature of the materials and the research question, a starting point was that qualitative analysis using human annotation was likely to be more valid and efficient than, for example, an LLM approach. While previous work has highlighted the promise of using LLMs to classify text \citep{Heseltine2024, Zhang2023, Gilardi2023}, there are also limitations, such as lack of consistency in factual queries \citep{Elazar2021, Hagstrom2023} and susceptibility to being misled by (invalid) arguments \citep{Wang2023}. Most importantly, LLMs tend to have a general lack of understanding of the underlying concept \citep{West2024}. Since the underlying concepts and arguments are crucial for our study, we expected that human judgement should be at the centre of the evaluation of the textual material \citep{Ronningstad2024, Tornberg2024, Kim2024}.\footnote{We also briefly experimented with an LLM approach (using the LLM Gemma) for this study, but found that the risk classification was not reliable.} We carried out a qualitative content analysis of the debate material, based on a systematic reading of the texts. 

For our coding exercise, we had the texts read and classified by three academics who are native Swedish speakers, have a PhD and come from three different disciplines: economics, computer science and political science.

For our data collection we used Mediearkivet\footnote{We have collected the articles via \url{https://app.retriever-info.com/services/archive}. Due to licensing issues, the articles cannot be freely shared. A full list of the articles included here, including the title of the article, the name(s) of the author(s) and the date of publication, is available on request.}, where we searched for the terms ``artificiell intelligens'' or ``AI'' or ``artificial intelligence'' in articles placed in either the debate or opinion section between 01.01.2000 and 25.10.23. This resulted in 318 articles (after removing duplicates and misidentified pieces). Almost all of them are a single page, although the actual length of the text varies. Note that this results in a selection of articles that are both about AI and those that only mention AI as a simple example. For example, the article ``Koppla ihop militär och civil forskning'' only mentions artificial intelligence as an example of one type of research (``Areas such as cyber security, communication, robotics and artificial intelligence are crucial to the defence of Sweden''\footnote{``Områden som cybersäkerhet, kommunikation, robotik och artificiell intelligens är helt avgörande för försvaret av Sverige''}). We then looked at all the articles individually and judged whether they were really ``about'' AI or just mentioned it in passing. This led to a significant reduction to 74 articles.

All texts were downloaded in their PDF format, cropped to show only the article in question (author names and affiliations were removed), and converted to .txt format using \texttt{pdftools} in R. These latter files were then checked and corrected manually (for example, to deal with strange characters, font size issues, and problems with columns). Each article was then split into four sections: title, header, body and additional notes. Of these, only the body is included in the later analysis. Metadata was also collected for each article, including a unique document ID, publication date, newspaper title, article title, section topic (if mentioned), additional notes, authors and affiliations.

For all articles, we also note the authors (of which there were often more than two) and their affiliations. We then reduce these affiliations to four broad categories: corporate (related to companies and businesses), NGO (all non-governmental and non-commercial organisations), political (related to political parties or holding official positions) and academic (related to universities and research institutes), as well as a combination of those working for companies and those working in academia. Of these categories, most were written by those associated with academia (31 articles), followed by companies (15 articles), politicians (14 articles), NGOs (10 articles) and the combination of companies and academia (4 articles).

We divided the texts between the three human coders, who were given the following instructions:

\begin{enumerate}
    \item Give a single sentence summary of what the text is about
    \item Give a label for the text in a single word (or two if its a common expression such as European Union)
    \item Indicate the risk on a scale of three categories: a) no risk, b) short-term risk, c) long-term risk
    \item Say why the text was coded into that risk category

\end{enumerate}

To be classified in one of the risk categories, the concept of risk had to be part of the core argument of the article, not just mentioned in passing. A short-term risk classification concerned a risk related to a concrete problem or phenomenon here and now, whereas a long-term risk concerned hypothetical scenarios and/or existential threats to humanity. Given our materials, it is difficult to establish strict definitions of short-term vs. long-term risk, along the lines of, for example, the World Economic Forum's distinction between 2-year and 10-year risks \citep{WorldEconomicForum2023a}. Authors of argumentative texts are rarely as specific in their writing, so we rely on our overall assessment of the argument. Since we have three different coders involved in reading the texts, the validity should increase. In our analysis below, we give examples of the current risk classification. 

\section{Results}

As mentioned in the methodology section, we coded a total of 74 articles. We will first look at the categories given by the human coders to better understand what the articles we selected were about. We then turn to the coding of risk, looking at the evolution of risk focus over time. Finally, we look at the authorship of the articles to see if certain authors preferred to talk about different types of risk - or not.

\subsection{Categories}

\begin{table*}
\caption{Overview of the labels and label counts as assigned by the coders. The Group labels are based on the 36 labels given by the coders.}
\label{tab:risk_cat_coders}
\begin{tabular}{llp{8cm}}
\toprule
\textbf{Group}           & \textbf{Count} & \textbf{Example Summary}  \\
\midrule
Ethics & 16 & There is a risk in believing that AI can assist decision-making in moral dilemmas \\
Regulation & 10 & Criticism of EU AI regulation\\
Risk & 9 & AI is developing too fast, ChatGPT is just a small step from AGI, this is a major existential risk for humanity \\
Applications & 7 & AI could help us prevent and solve crime\\
Characteristics & 6 & This is what AI is and could be in the future\\
Economy & 6 & The author argues that AI will transform the retail sector\\
Labour market & 5 & Young people will find it harder to get their first job as simpler tasks are automated by AI systems\\
Development & 5 & Suggests the creation of a committee to deal with AI issues \\
Education & 3 & The author argues that people need to be better educated about the benefits and risks of AI \\
Research & 3 & More interdisciplinary AI research is needed \\
Sustainability & 2 & The author argues that we need to invest in sustainable AI by integrating responsible and transparent AI practices \\
Techno-optimism & 2 & Technological change should be accelerated to benefit the economy, fears of mass unemployment are unfounded \\
\bottomrule
\end{tabular}
\end{table*}

Table \ref{tab:risk_cat_coders} shows the labels assigned to each article by the coders, as well as an example in the form of a one-sentence description of an article of that type. The most common label is ``Ethics''. These articles describe various topics related to the ethical issues surrounding AI. Sometimes these arguments include ``warnings'' about the potential threat that AI can pose to democracy, or the biases that can arise from using AI trained on poorly curated data, and the lack of transparency associated with this issue. Other issues discussed include the inherent limitations of an AI ever becoming human, such as the impossibility of having a moral compass or ever developing conscious awareness. Although this is related to risk, not all of them mention it, and when they do, it is mostly related to the idea of short-term risk.

The second category, Regulation, deals with various calls for the regulation of AI, such as criticism of the European Union's proposed Artificial Intelligence Act, calls for harmonisation of AI regulation between the US and the EU, and the risks posed by the lack of regulation. This category is related to, but different from, the Development category in that the latter is more focused on calls to action, focusing on how AI should be managed and developed in the country. The third most common category - risk - discusses issues such as the risks of AI being used for autonomous weapons, the impact of unreliable and biased training data, and concerns about integrity. Of the 9 articles in this category, 6 are categorised as dealing specifically with long-term risks. These mostly concern the idea that current AI development is too fast and that future systems could easily lead to Artificial General Intelligence (AGI).

The fourth and fifth categories relate to descriptions of what \emph{what} AI is and various applications in which it might be used. Here we find examples of how AI can be used by the police to help solve crimes, or how hospitals can use it to digitise many of their systems, or how it can help reduce the administrative burden in public institutions. The sixth and seventh categories are related, but while the ``economy'' category focuses on the economic sector (most often commerce) in general, the ``labour market'' category focuses mostly on the impact of AI adoption on people's jobs. In the latter category, while most authors agree that AI will transform the labour market, the disagreement lies in whether this is seen as a positive development (for example, that AI will create more jobs and opportunities) or whether this development is negative (for example, in relation to the first jobs young people take).

As for the other categories, ``Education'' focuses on the development of skills related to AI as well as further investment in more flexible education systems to cope with rapidly changing markets, ``Research'' includes articles calling for more (interdisciplinary) research, ``Sustainability'' focuses on different ways to ensure that AI and its developments will be sustainable in the future, while ``Techno-optimism'' includes two articles that mainly push the framework that any technological change is beneficial and should be supported and accelerated.

\subsection{Types of Risk}

\begin{figure}[!ht]
    \centering
    \includegraphics[width=\linewidth]{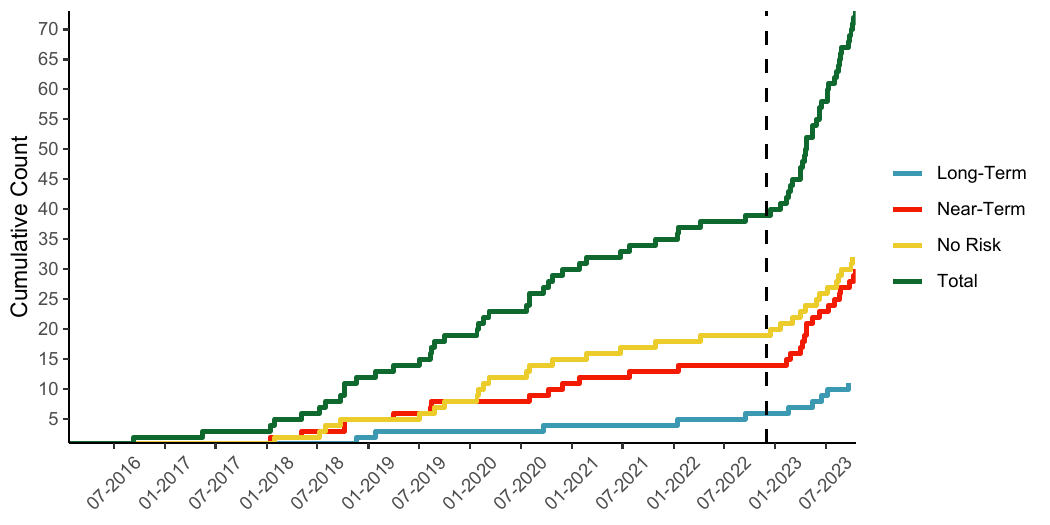}
    \caption{Cumulative Number of Articles over time classified by the coders in the three different types of risk. The vertical dashed line indicates the release of ChatGPT on November 30, 2022}
    \label{fig:risk_time}
\end{figure}

Figure \ref{fig:risk_time} illustrates the evolution of the assigned risk categories over time, as determined by the coders. It is noteworthy that from January 2023, which coincides with the release of ChatGPT, there was a significant increase in the number of articles published. This increase was most pronounced for short-term risk articles, while long-term risk articles showed a more modest increase. Overall, the majority of articles did not contain a core argument about risk. An example of long-term risk comes form an article accusing ``leading AI companies [to play] Russian roulette''\footnote{Göteborgs Posten, 18-02-2023}, with the article making the argument that AI development in general is moving too fast and that techniques such as ChatGPT are only a small step away from Artificial General Intelligence (AGI) and therefore a major existential risk to humanity. An example of short-term risk comes from an article a about the current use of AI by law enforcement and how it is unreliable (for example, when used for facial recognition) due to racial bias in the training data\footnote{Sydsvenskan, 31-07-2020}. Finally, an example of no risk comes from an article entitled ``AI is redrawing the labour market and our skills needs\footnote{Dagens industri, 15-12-2022}, which does not identify any risks at all, but instead argues that Sweden needs to invest in the data-driven economy in order not to fall behind the AI development of other countries.

\subsection{Authorship}

\begin{table*}
\small
\label{tab:risk_cat_table}
\caption{Classification of Risk per Affiliation type}
\begin{tabular}{lllllll}
\toprule
     & \textbf{Academic}      & \textbf{Industry} & \multicolumn{1}{p{1.8cm}}{\textbf{Industry-Academic}} & \textbf{NGO} & \textbf{Political} &  \textbf{Total} \\
\midrule
Long-Term             & 8             & 1       & 2                & --  &   --        &    11         \\
Short-Term             & 14            & 6       & 3                & 5   & 3         & 31          \\
No Risk               & 9             & 9       & 1                & 4   & 9         & 32          \\
&&&&&& \\
Total      & 31            & 15      & 4                & 10  & 14        & 74          \\ 
\bottomrule
\end{tabular}
\end{table*}

As Table \ref{tab:risk_cat_table} shows, most of the articles (31 out of 74) were written by authors associated with the academic community, such as professors or researchers, followed by those associated with industry - most often working in the field of data or computer science; politicians, either speaking on behalf of their party or on behalf of the government; and representatives of NGOs. In most cases, articles were written by one or more authors from a single group, with the only exception being collaborations between those working in industry and those working in academia. Looking at how the different types of authors wrote about risk, we find that those associated with NGOs and politicians never wrote about long-term risk. Instead, these two types either talk equally about short-term risk, or do not mention risk at all, or have a clear preference for the latter, as is the case with politicians. Thus, during the studied period, politicians tended to focus their pieces on neutral aspects of AI rather than highlighting risk.

\begin{table*}
\label{tab:cat_academic}
\caption{Classification of Short-Term and Long-Term Risk per Academic discipline}
\begin{tabular}{p{4cm}llll}
\toprule
      & \textbf{Long-Term}            & \textbf{Short-Term} & \textbf{No Risk} & \textbf{Total} \\
      \midrule
Business Administration   & --                         & 3             & 2           & 5                    \\
Chemistry                 & --                         & --              & 1           & 1                    \\
Computer Science          & --                         & 13            & 2           & 15                   \\
Data Science              & 1                        & --              & 1           & 2                    \\
Economics                 & 3                        & 1             & 2           & 6                    \\
Linguistics               & --                         & 1             & --            & 1                    \\
Mathematics               & 4                        & 1             & --            & 5                    \\
Media Science             & --                         & 1             &  --           & 1                    \\
Neuroscience              & 2                        & --              & --            & 2                    \\
Philosophy                & 3                        & 1             & 1           & 5                    \\
Physics                   & 1                        & --              &   --          & 1                    \\
Social Science            & --                         & 8             & 1           & 9                    \\
\bottomrule
\end{tabular}
\end{table*}

Given that so many of the long-term risk articles are written by those who belong to academia, it might be interesting to look at the type of academic discipline they belong to, as table \ref{tab:cat_academic} shows. It is interesting to note that while computer science is the most common category, none of the authors here wrote an article in the long-term risk category, opting for short-term risk or not mentioning risk at all. This suggests that, of all the academic disciplines, computer scientists in the Swedish context do not want to focus on the long-term aspects. Instead, articles on long-term risk were most often written by someone working in mathematics, economics or philosophy. In fact, all four articles written by a mathematician was by the same author. This suggests that a few academics with experience in different academic fields are highlighting existential risks, while the majority are not. Another finding is that AI is not (yet) 'owned' by politicians, but rather by an academic elite.

\section{Conclusions}

This study analysed the evolution of the AI debate in a national context, based on a human reading of the argumentative texts to identify the evolution over time. We began our study with the expectation that the character of the AI debate has changed over the past few years, and that this should be a top-down development. As the dataset contains arguments and is relatively small, we expected that human expert coding would be essential, as opposed to categorising pure political news material. We therefore conducted a qualitative content analysis. 

Our results indicate that the emphasis on short-term risks has increased in the Swedish public debate since the release of ChatGPT. This finding complements previous analyses of the prevalence of risks in the news media, for example \citet{Nguyen2023a}, as we show that the focus on short-term risks has increased since 2022. Furthermore, our analysis supports the expectation that academics rather than politicians will take the lead in this debate. This signals a weakening of the parties in Sweden relative to other arenas of society in relation to previous patterns in this context, e.g. \citet{Holmberg1997a}. Instead, it seems that it is in the supranational EU context that important directions of development take place. 

Another reflection is that the Swedish AI debate seems to be somewhat more nuanced than, for example, the debate in Canada, where \citet{Dandurandetal2023} argue that the media tend to cool down controversies and risks. The conclusions drawn will also depend on the material used for the study. In the present study, we focused on argumentative texts rather than general news coverage. Another finding is that the long term risk perspective was not very present in our sample, compared to what seem to be the case among AI developers and computer scientists in the US. In the Swedish context, the overall debate focuses more on short term risks or is neutral, especially if a politician takes the tone. 

A relevant follow-up to our study would be to analyse the AI debate in other national EU contexts: neighbouring countries such as Norway and Denmark, or more contrasting cases such as Spain, in order to be able to generalise our findings to a greater extent. EU regulation is still in its infancy, and country differences are likely to continue to play an important role.

\subsection*{Acknowledgments}
This work was supported by the Marianne and Marcus Wallenberg Foundation and the Wallenberg AI, Autonomous Systems and Software Program – Humanity and Society. Bastiaan Bruinsma is grateful for support from the GATE Project, funded by the European Union's Horizon 2020 WIDESPREAD-2018-2020 TEAMING Phase 2 programme under Grant Agreement No. 857155.

\subsection*{Declaration of Interest}

One of the authors, Moa Johansson, declares to have co-authored two opinion articles analysed here - ``Skrämsel om AI döljer de verkliga problemen'', published in Göteborgs Posten on 24-02-2023, and ``Alla måste lära sig mer om artificiell intelligens'', published in Göteborgs Tidningen on 02-04-2019. To avoid potential conflicts of interest, both articles were coded by one of the other two coders. 

\bibliography{lib_debatt}

\end{document}